\documentclass[preprint,12pt]{elsarticle}

\usepackage{amssymb}

\usepackage{amsmath}
\usepackage{algorithm}
\usepackage{algpseudocode}
\usepackage{booktabs}
\usepackage{multirow}
\usepackage{url}
\usepackage{makecell}
\usepackage{pgfplots}
\pgfplotsset{compat=1.18}
\usepackage{placeins}

\journal{Information Fusion}

\begin{document}

\begin{frontmatter}

\title{CSG-Nav: Confidence-Aware Semantic-Geometric Fusion for Robust Exploration and Accurate Zero-Shot Object-Goal Navigation}

\author[label1,label2]{Shoubin Chen} 
\author[label1,label2]{Rui Xu}
\author[label1,label2]{Zehao Wu}
\author[label1,label2]{Kai Zhang\corref{cor1}}
\ead{kaizhangpostbox@hotmail.com}
\cortext[cor1]{Corresponding author}
\author[label1]{Tao Huang}
\author[label1,label2]{Zhou Liu}
\author[label1,label2]{Qingquan Li}
\affiliation[label1]{organization={Guangdong Laboratory of Artificial Intelligence and Digital Economy (SZ)},
            city={Shenzhen},
            postcode={518132}, 
            state={Guangdong},
            country={China}}

\affiliation[label2]{organization={Shenzhen University},
            city={Shenzhen},
            postcode={518061}, 
            state={Guangdong},
            country={China}}

\begin{abstract}
Zero-shot object-goal navigation requires autonomous agents to explore unseen environments and locate target objects without task-specific training, posing challenges in perceptual reliability under observation uncertainty and physically grounded goal termination. Although foundation-model-based approaches improve generalization, they remain vulnerable to unreliable long-range perception, hallucination-induced directional bias, and semantically plausible but geometrically incorrect stopping decisions. To address these limitations, we propose CSG-Nav, a zero-shot navigation framework centered on confidence-aware multimodal information fusion. CSG-Nav integrates three complementary fusion mechanisms. First, semantic-source fusion integrates open-vocabulary detection results, detection confidence, scene observations, and relational reasoning to produce more reliable direction-level semantic scores and suppress hallucination-induced directional bias. Second, spatial-confidence fusion weights multi-view semantic evidence using a Gaussian decay function, suppressing unreliable distant observations while accumulating spatially consistent exploration preferences. Third, semantic-geometric decision fusion combines semantic target detection with depth-based 3D localization to support physically grounded stopping decisions. Experiments on HM3Dv1, HM3Dv2, and MP3D demonstrate competitive zero-shot performance, achieving success rates of 58.9\%, 71.6\%, and 44.3\%. Fusion-level ablations and real-world deployments in indoor and outdoor environments further verify the effectiveness and practicality of CSG-Nav for robust navigation under uncertain perception. Code is available at \url{https://xr-ui.github.io/CSGNav.github.io/}.
\end{abstract}



\end{frontmatter}



\section{Introduction}
Object-goal navigation is a fundamental capability for autonomous robots operating in human-centered environments, where an agent must explore an unseen environment and locate a target object specified by a natural-language command, such as ``find a sofa''. Training-based approaches have achieved significant progress in this task~\citep{LI2026104379}, but their reliance on extensive task-specific training limits scalability and generalization in open-world environments. Zero-shot object navigation has therefore emerged as a promising paradigm, requiring agents to navigate without task-specific training. In practical robotic systems, however, zero-shot navigation is not merely a semantic recognition problem; it is also an information fusion problem. The agent must integrate heterogeneous evidence from object detectors, vision-language observations, depth sensors, spatial maps, and sequential multi-view observations to make reliable decisions under partial observability, noisy perception, and uncertain spatial evidence.

Recent foundation-model-based methods have improved open-vocabulary generalization by using pretrained vision-language models (VLMs) and large language models (LLMs). Early methods, such as ZSON~\citep{majumdar2022zson} and CoW~\citep{gadre2023cows}, employ pretrained VLMs such as CLIP~\citep{radford2021learning} for open-vocabulary object localization, while VLFM~\citep{yokoyama2024vlfm} projects VLM-based semantic similarity scores onto a top-down value map to guide exploration. Other approaches, including L3MVN~\citep{Yu_2023} and ESC~\citep{zhou2023esc}, leverage LLM-based commonsense reasoning to infer object-scene relationships and prioritize candidate search regions. More recently, WMNav~\citep{11246684} uses VLMs as a world-model component to predict future observations and reduce ineffective exploration. Despite these advances, many existing methods still propagate semantic outputs directly into planning without sufficiently modeling how reliable, spatially consistent, or physically grounded these outputs are. Object detectors provide object-level evidence, including categories, bounding boxes, and confidence scores. 
VLM observations provide complementary scene-level context. 
The key challenge is how to fuse these heterogeneous sources with reliability modeling, multi-view consistency, and geometric reachability for robust navigation decisions.

\begin{figure*}[ht]
\centering
\includegraphics[width=0.9\textwidth]{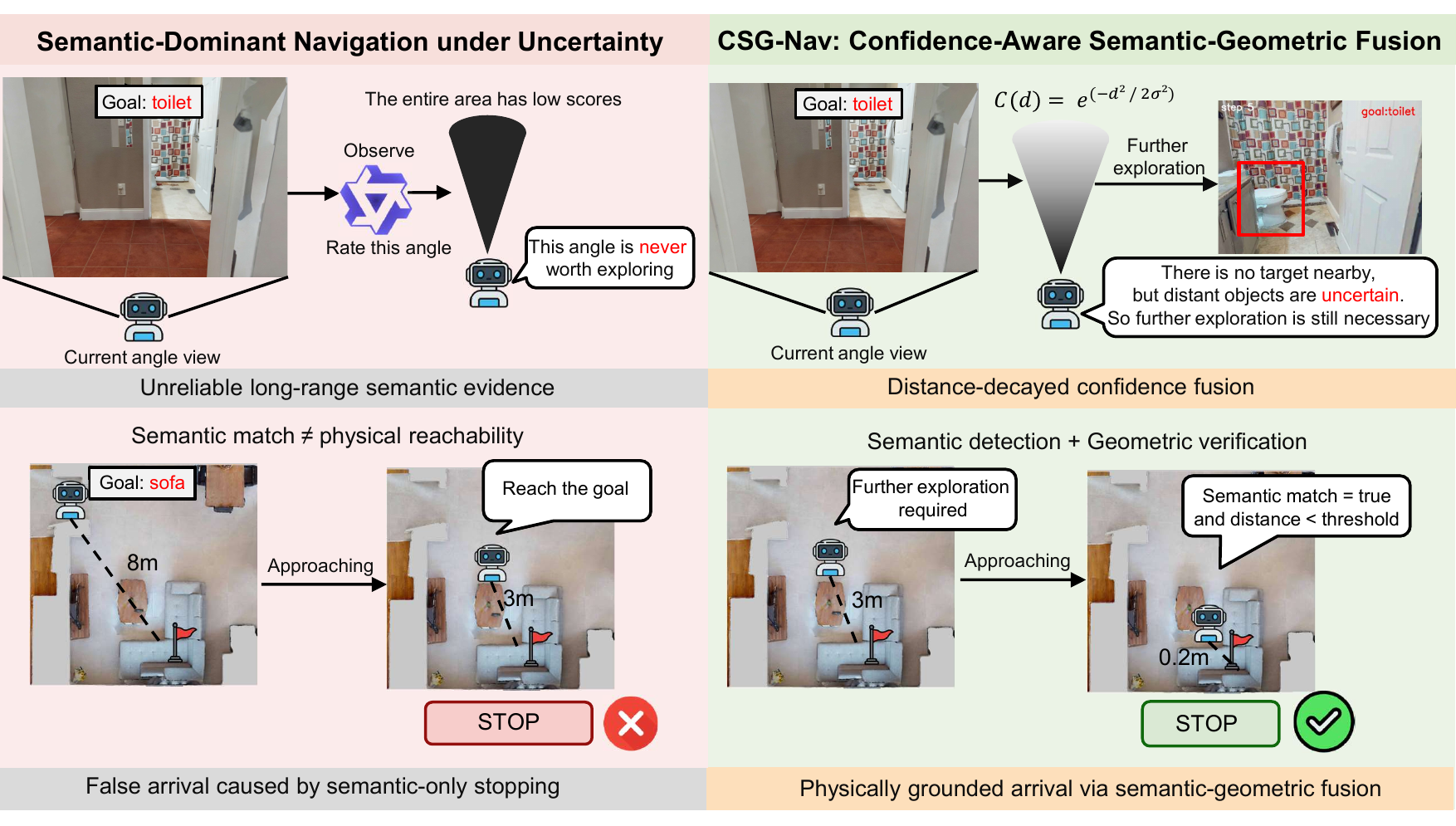}

\caption{Motivation of CSG-Nav from an information fusion perspective. Existing semantic-dominant zero-shot navigation methods may suffer from unreliable long-range semantic evidence and false arrival caused by semantic-only stopping. CSG-Nav addresses these issues through confidence-aware semantic-spatial fusion for robust exploration and semantic-geometric decision fusion for physically grounded arrival verification.}
\label{fig:motivation}
\end{figure*}

From an information fusion perspective, three critical gaps remain insufficiently addressed. First, there is a lack of context-aware information mediation between perception and reasoning. When semantic scores generated by VLMs are treated as direct control signals, hallucinated or weakly supported cues can propagate through the decision chain and bias navigation toward semantically plausible but unreliable directions. Second, the spatial degradation of multi-view visual evidence is often ignored. According to sensor characteristics, distant observations usually suffer from reduced image resolution, lower object-detection confidence, and noisier depth estimates; consequently, global semantic projection without distance-aware reliability modeling can contaminate the preference map with low-confidence long-range evidence, as shown in the top panels of Fig.~\ref{fig:motivation}. Third, semantic target detection is often treated as sufficient evidence for arrival. 
However, as illustrated in the bottom panels of Fig.~\ref{fig:motivation}, visual recognition does not guarantee geometric reachability or a valid stopping pose. This distinction is especially important for downstream manipulation and service tasks, where the robot must reach a physically meaningful position before interacting with the target object. These limitations motivate a confidence-aware fusion mechanism that can integrate semantic, spatial, and geometric evidence while suppressing unreliable observations.

To address these challenges, we propose CSG-Nav, a confidence-aware semantic-geometric fusion framework for robust zero-shot object-goal navigation. Rather than treating foundation-model outputs as isolated decision signals, CSG-Nav performs multimodal information fusion at three levels. First, semantic-source fusion integrates open-vocabulary detection results, detection confidence, VLM observations, and structured LLM reasoning to produce more reliable semantic evidence and reduce hallucination-induced directional bias. Second, spatial-confidence fusion models the reliability of multi-view observations using a Gaussian decay function, allowing the agent to suppress unreliable distant evidence while accumulating spatially consistent exploration preferences. Third, semantic-geometric decision fusion combines semantic target detection with depth-based 3D localization to verify whether the target is physically reachable before issuing the stopping decision. Through these fusion mechanisms, CSG-Nav improves both exploration robustness and goal-termination reliability under uncertain perception.

Our contributions can be summarized as follows:
\begin{enumerate}

\item \textbf{Confidence-Aware Multimodal Navigation Framework:} 
We propose CSG-Nav, a zero-shot object-goal navigation framework that integrates semantic evidence, spatial confidence, and RGB-D geometry.

\item \textbf{Semantic-Source Fusion:} 
We integrate object detections and detection confidence with VLM scene observations through structured LLM reasoning to obtain more reliable direction-level semantic scores.

\item \textbf{Spatial-Confidence Fusion:} 
We apply Gaussian distance-conditioned reliability weighting to reduce the influence of distant observations and build a more robust global preference map.

\item \textbf{Semantic-Geometric Decision Fusion:} 
We fuse target detection with depth-based 3D localization to avoid premature stopping when the target is semantically visible but has not yet satisfied the geometric arrival condition.

\item \textbf{Fusion-Oriented Evaluation:} 
We validate CSG-Nav on HM3Dv1, HM3Dv2, MP3D, and real-world scenarios with fusion-level ablations and failure analysis.

\end{enumerate}

The remainder of this paper is organized as follows. Section~\ref{sec:related_works} reviews zero-shot object-goal navigation, foundation models for navigation, and multimodal information fusion in navigation. Section~\ref{sec:method} details the proposed CSG-Nav framework from a fusion-oriented perspective, Sections~\ref{sec:simulated_experiments} and~\ref{sec:real-world_experiments} present experimental evaluations in simulated and real-world environments, respectively. Finally, Section~\ref{sec:conclusions} concludes the paper and discusses limitations and future research directions.

\section{Related Work}
\label{sec:related_works}

\subsection{Zero-shot Object Goal Navigation}
Zero-shot object-goal navigation requires an agent to locate a target object in an unseen environment without task-specific training. Early methods such as ZSON~\citep{majumdar2022zson} and CoW~\citep{gadre2023cows} use pretrained VLMs for open-vocabulary object localization, while VLFM~\citep{yokoyama2024vlfm} projects semantic similarity  onto a spatial value map. Subsequent studies introduce richer semantic reasoning and exploration strategies. ESC~\citep{zhou2023esc} uses LLM commonsense to reason about target-relevant regions. Related learning-based object-navigation studies have also investigated task decomposition. Multiple Thinking~\citep{dang2023multiple} separates object navigation into specialized meta-abilities, whereas Dual Adaptive Thinking~\citep{dang2023search} distinguishes target search from post-detection navigation. Recent methods further investigate unified goal representations, episodic memory, and adaptive exploration, including UniGoal~\citep{yin2025unigoal}, ReMemNav~\citep{wu2026rememnav}, curiosity-driven navigation~\citep{gu2026curiosity}, and Stairway to Success~\citep{gong2026stairway}.

Across these lines of research, existing methods have advanced semantic reasoning, goal representation, memory, exploration, and task decomposition from different perspectives. However, the reliability of heterogeneous evidence across semantic perception, spatial accumulation, and stopping decisions has received comparatively less attention. CSG-Nav focuses on this aspect by combining complementary semantic cues, confidence-aware spatial evidence accumulation, and geometric verification within a unified navigation framework.

\subsection{Foundation Models for Navigation}

Foundation models have been widely used in embodied navigation as semantic scorers, commonsense reasoners, and world models. As semantic scorers, VLMs estimate the relevance between observations and navigation goals. CoW~\citep{gadre2023cows} selects target-related directions using VLM cues, VLMaps~\citep{huang2023visual} fuses vision-language features with spatial maps for language-conditioned search, VLFM~\citep{yokoyama2024vlfm} builds a semantic value map from VLM scores, and TopV-Nav~\citep{zhong2025topvnavunlockingtopviewspatial} performs reasoning over top-view spatial representations.

As commonsense reasoners, LLMs provide object-scene and object-room priors for high-level planning. L3MVN~\citep{Yu_2023} and ESC~\citep{zhou2023esc} use LLM commonsense to infer likely target regions, while LFG~\citep{shah2023navigation} incorporates LLM-generated heuristics into navigation. LROGNav~\citep{sun2025enhancing} further injects LLM-derived room-object relations into a modular map-based pipeline and uses multimodal fusion to improve search efficiency.

A third line of work explores predictive world models and generative policies for embodied navigation and robot decision-making. NWM~\citep{bar2025navigation} uses video generation to predict future visual observations for navigation planning, while Unified World Models~\citep{zhu2025unified} jointly models visual dynamics and robot actions through video and action diffusion. For object goal navigation, WMNav~\citep{11246684} employs VLM-based world modeling with curiosity memory to preserve predicted environmental states and support subsequent planning. Related generative navigation policies, such as FLUX~\citep{gong2026flux}, use rectified flow and static-to-dynamic learning to generate navigation actions across different robot embodiments. More broadly, recent work on agentic world modeling~\citep{chu2026agentic} discusses predictive environment representations as a general mechanism for supporting interactive agents. These predictive and generative approaches mainly improve navigation through future-state prediction or direct action generation. In contrast, CSG-Nav focuses on reliability-aware fusion of currently available semantic, spatial, and geometric evidence.

\subsection{Multi-Source Information Fusion in Navigation}

Multi-source information fusion is central to embodied navigation because agents must integrate visual observations, language goals, spatial representations, and historical information for action selection. Vision-and-language navigation (VLN) provides a representative example of this problem. A recent survey~\citep{wang2026comprehensive} organizes VLN methods according to their action and model paradigms and emphasizes the joint use of visual observations, language instructions, spatial context, and navigation history for decision-making. OmniNav~\citep{xue2026omninav} further unifies instruction-goal, object-goal, and point-goal navigation with frontier-based exploration through a fast-slow architecture. These studies demonstrate the importance of integrating visual, linguistic, spatial, and historical information. Building on this broader perspective, CSG-Nav further emphasizes the reliability of semantic, spatial, and geometric evidence during zero-shot object-goal navigation.

Map-based methods accumulate semantic observations into spatial representations to support exploration~\citep{chaplot2020object,yokoyama2024vlfm,kuang2024openfmnav}. Recent approaches further incorporate uncertainty into these representations. Uncertainty-Aware Gaussian Map~\citep{gao2026uncertainty} estimates geometric, semantic, and appearance uncertainty within a semantic Gaussian map and incorporates them into a unified 3D value map. Beyond navigation, uncertainty-aware 3D reconstruction~\citep{liu2026uncertainty} models observation-dependent uncertainty to suppress unreliable measurements in dynamic underwater scenes. 

Unlike learned or probabilistic uncertainty-estimation methods, CSG-Nav does not aim to model all sources of perceptual uncertainty. Instead, it focuses on distance as an explicit and readily available reliability factor during spatial evidence fusion, since observation quality generally degrades with increasing sensing range. A training-free Gaussian distance prior is therefore used to reduce the influence of distant observations.

\section{System Design}
\label{sec:method}

\subsection{Overview of Multimodal Fusion Architecture}
\begin{figure*}[ht]
\centering
\includegraphics[width=0.9\textwidth]{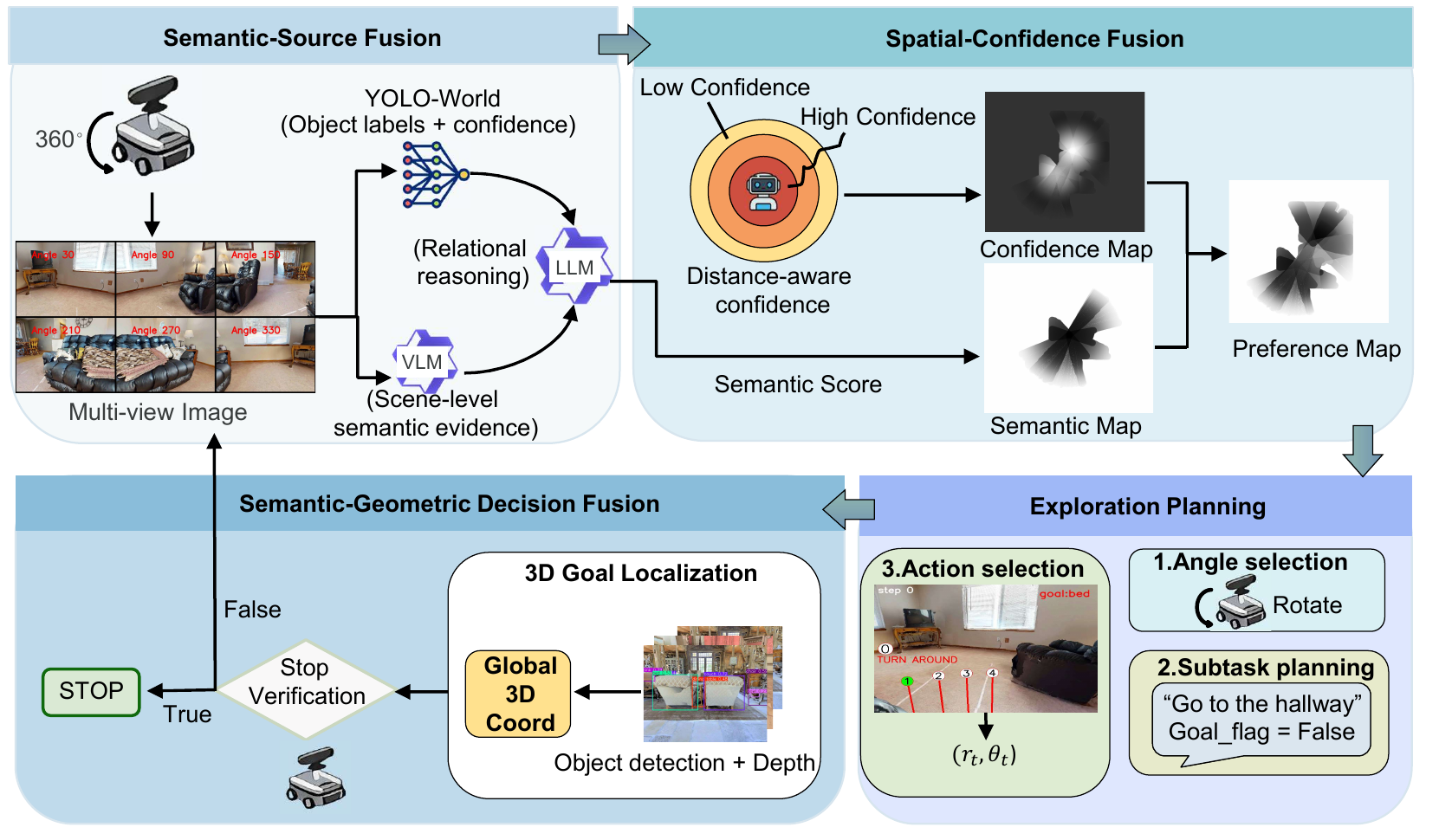}

\caption{Overview of the CSG-Nav multimodal fusion architecture. The framework integrates semantic-source fusion, spatial-confidence fusion, and semantic-geometric decision fusion to support exploration planning and goal-termination verification.}

\label{fig:overview}
\end{figure*}

We propose CSG-Nav, a confidence-aware multimodal fusion framework for zero-shot object-goal navigation, as shown in Fig.~\ref{fig:overview}. Rather than treating perception, mapping, planning, and verification as independent stages, CSG-Nav integrates heterogeneous evidence from open-vocabulary object detection, vision-language scene understanding, structured language reasoning, multi-view spatial observations, accessibility estimation, and RGB-D geometry.

At each navigation step, the agent first acquires multi-view RGB-D observations. Semantic-source fusion integrates object-level and scene-level evidence to produce direction-level semantic scores. Spatial-confidence fusion then combines these scores with distance-conditioned reliability and accessibility constraints to update a global preference map for exploration. When the target is detected, semantic-geometric decision fusion combines target detection with depth-based 3D localization to determine whether the geometric arrival condition is satisfied before issuing the stop decision. These three complementary fusion mechanisms operate at different stages of the navigation pipeline and jointly constitute the multi-level fusion framework of CSG-Nav. Table \ref{tab:fusion_hierarchy} summarizes the three fusion levels of CSG-Nav, including their input sources and corresponding outputs.

\begin{table}[htbp]
\centering
\caption{Fusion hierarchy of CSG-Nav.}
\label{tab:fusion_hierarchy}
\begin{tabular}{p{0.20\linewidth} p{0.48\linewidth} p{0.20\linewidth}}
\hline
\textbf{Fusion level} & \textbf{Sources fused} & \textbf{Output} \\
\hline
Semantic-source fusion
& YOLO-World detections, detection confidence, VLM scene descriptions, and LLM commonsense reasoning
& Directional semantic scores \\
\hline
Spatial-confidence fusion
& Multi-view semantic scores, Gaussian distance-conditioned reliability, and accessibility constraints
& Global preference map \\
\hline
Semantic-geometric fusion
& Semantic target detection, depth-based 3D localization, and distance threshold
& Stop / continue decision \\
\hline
\end{tabular}
\end{table}

\subsection{Semantic-Source Fusion Module}

\begin{figure*}[ht]
\centering
\includegraphics[width=0.9\textwidth]{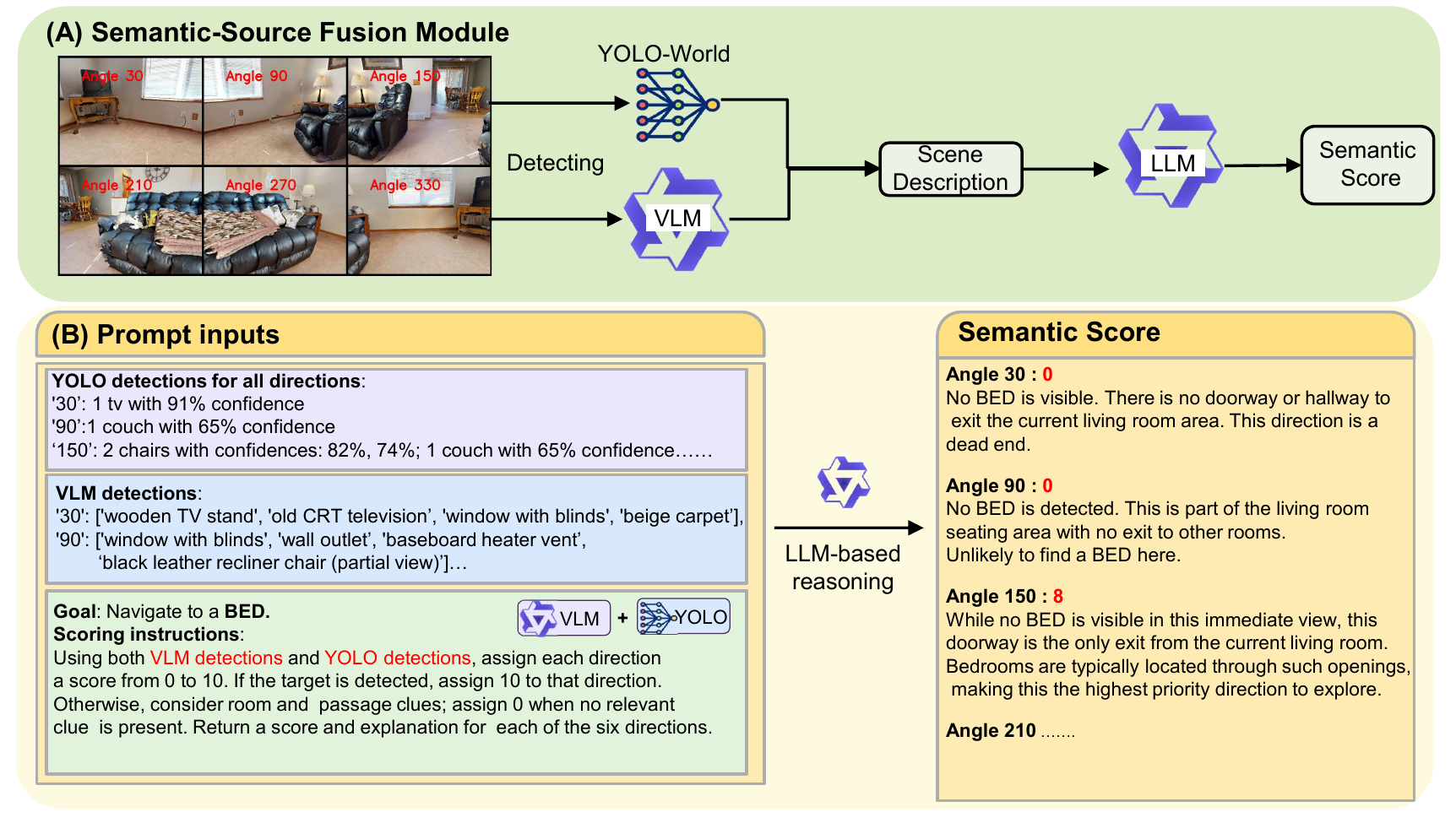}
\caption{Semantic-source fusion in CSG-Nav. For each candidate direction, object-level detection results and scene evidence are fused through structured LLM reasoning to produce a semantic score and rationale.}
\label{fig:perception}
\end{figure*}

At the beginning of each step, the agent scans six candidate angles, $\alpha \in \{30^\circ,90^\circ,150^\circ,210^\circ,270^\circ,330^\circ\}$, and obtains the corresponding RGB-D observations $I_\alpha$. The six RGB views are concatenated into a panoramic observation and processed by the VLM to obtain scene-level observations for each direction. In parallel, YOLO-World~\citep{cheng2024yolo} processes the directional images individually to extract object categories, bounding boxes, and detection confidence scores.

As shown in Fig.~\ref{fig:perception}, semantic-source fusion integrates the complementary object-level and scene-level evidence before navigation scoring. For each direction, the YOLO-World detections and confidence scores are converted into direction-indexed textual descriptions $D_{\alpha}^{\mathrm{det}}$ and combined with the corresponding VLM-generated scene observation $D_{\alpha}^{\mathrm{vlm}}$. Together with the target category and a predefined scoring instruction, these inputs are provided to the LLM, which jointly reasons over the two semantic sources to produce a direction-level semantic score and its accompanying rationale.

This semantic-source fusion can be written as:
\begin{equation}
\left(S_t(\alpha),R_t(\alpha)\right)=F_{\mathrm{sem}}\left(T,D_{\alpha}^{\mathrm{det}},D_{\alpha}^{\mathrm{vlm}},P_{\mathrm{score}}\right),
\end{equation}
where $T$ denotes the target category, and
$P_{\mathrm{score}}$ denotes the predefined scoring instruction. In the implementation, $F_{\mathrm{sem}}$ is realized through structured LLM reasoning over the two semantic sources.
The semantic score $S_t(\alpha)$ ranges from 0 to 10, where higher values indicate greater relevance to the target. The scoring instruction specifies a score of 10 when the fused semantic evidence indicates that the target object is detected. Otherwise, the LLM assigns scores from 0 to 9 according to the relevance of the detected objects, scene context, and their semantic relationship to the target. The accompanying rationale $R_t(\alpha)$ provides a textual explanation for the assigned score.

By formulating semantic perception as a multi-source fusion process, CSG-Nav avoids relying on a single semantic source  as the navigation signal. Object-level detections, detection confidence, and VLM scene context are jointly considered through LLM reasoning before influencing exploration. When the semantic sources provide inconsistent or weakly supported cues, the LLM evaluates them jointly under the same scoring instruction to produce the final direction-level score. This fusion process reduces the influence of isolated semantic errors and hallucination-induced directional bias, thereby improving the robustness of semantic scoring.

\subsection{Spatial-Confidence Fusion Module}
\label{sec:spatial-conf-fusion}

\begin{figure*}[ht]
\centering
\includegraphics[width=0.9\textwidth]{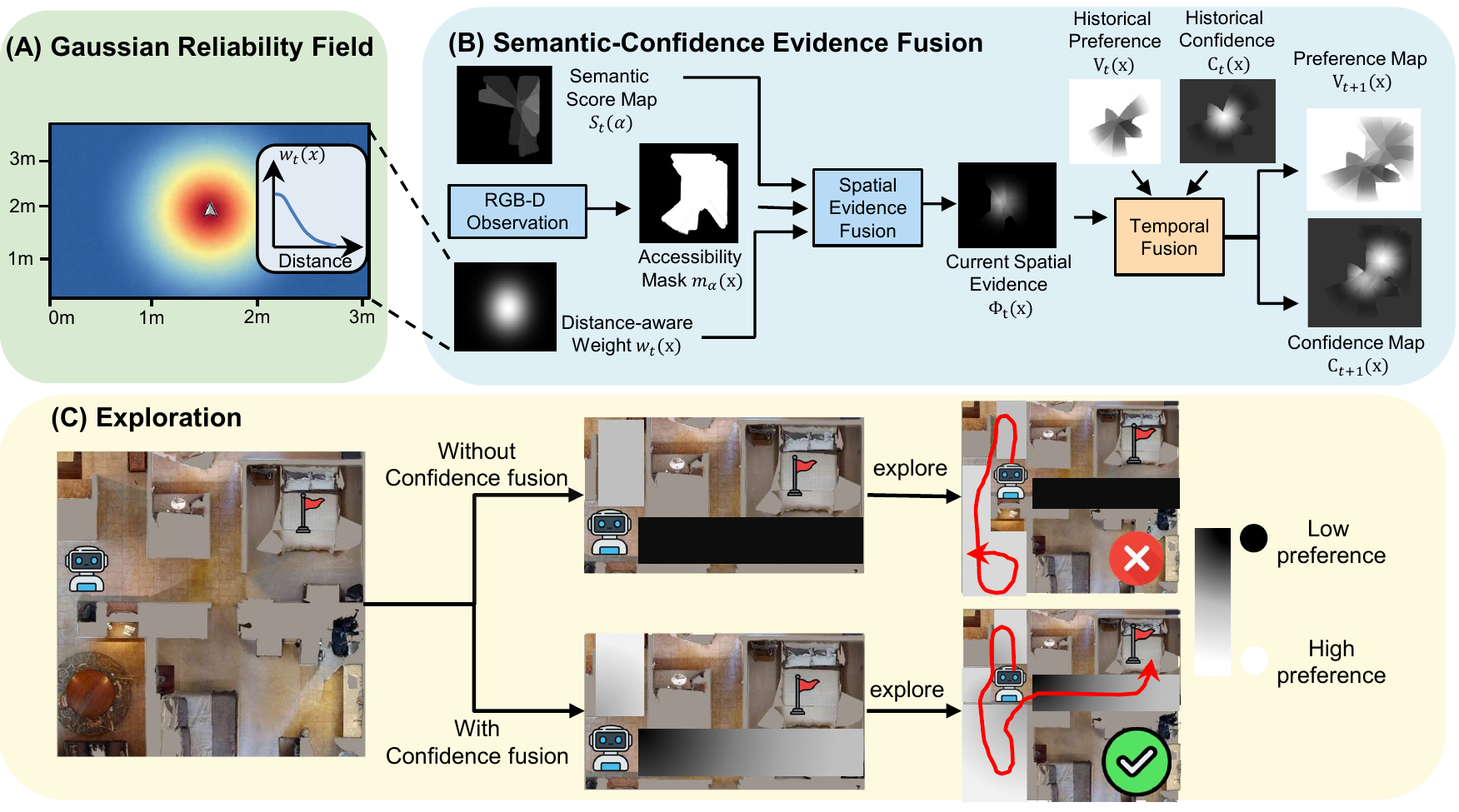}
\caption{Spatial-Confidence fusion. 
(A) Distance-aware reliability is modeled by a Gaussian decay function, where nearby observations receive higher reliability weights and distant observations receive lower weights. 
(B) Semantic scores are combined with accessibility and distance-aware reliability to form current spatial evidence, which is then integrated with historical preference and confidence information to update the global preference map.
(C) The bottom comparison shows that confidence-aware map updating suppresses unreliable distant evidence and produces more stable exploration preferences.}
\label{fig:preference_map}
\end{figure*}

After semantic-source fusion, CSG-Nav projects the direction-level semantic scores into a global preference map through spatial-confidence fusion, as illustrated in Fig.~\ref{fig:preference_map}. The module maintains a preference map $V(\mathbf{x})$, which represents target relevance or exploration priority at grid location $\mathbf{x}$, and a reliability map $C(\mathbf{x})$, which represents the fused confidence associated with the current preference estimate. In parallel, accessible regions are estimated from RGB-D observations and pose information to construct the mask $m_\alpha(x)$, which excludes obstructed or currently inaccessible areas from subsequent map updates.

As shown in Fig.~\ref{fig:preference_map}(A), the reliability of projected observations is modulated according to their distance from the agent. We define a Gaussian distance-conditioned reliability weight centered at the agent position $P_t$:
\begin{equation}
w_t(\mathbf{x})=\exp\left(-\frac{\|\mathbf{x}-P_t\|^2}{2\sigma^2}\right)
\end{equation}
where $\sigma$ controls the spatial decay scale. This weighting reflects the assumption that nearby observations are generally more reliable than distant ones and provides a simple, training-free reliability prior for spatial fusion.

As illustrated in Fig.~\ref{fig:preference_map}(B), the direction-level semantic score $S_t(\alpha)$ is projected onto the corresponding accessible region and modulated by the distance-conditioned reliability weight. The resulting current spatial evidence is defined as:
\begin{equation}
\Phi_t(\mathbf{x}) = m_{\alpha}(\mathbf{x}) \cdot w_t(\mathbf{x}) \cdot S_t(\alpha)
\end{equation}
where $m_{\alpha}(\mathbf{x})$ acts as an accessibility gate, $w_t(\mathbf{x})$ provides distance-aware reliability, and $S_t(\alpha)$ provides semantic relevance. This formulation reduces the influence of distant observations before they are incorporated into the global preference map.

The current spatial evidence is then integrated with the historical preference and reliability states through confidence-weighted temporal fusion. Following the confidence-weighted aggregation formulation in~\citep{yokoyama2024vlfm}, the preference map is updated as

\begin{equation}
V_{t+1}(\mathbf{x})=
\frac{C_t(\mathbf{x})V_t(\mathbf{x})+\Phi_t(\mathbf{x})}
{C_t(\mathbf{x})+m_{\alpha}(\mathbf{x})w_t(\mathbf{x})},
\quad \forall \mathbf{x}\in m_\alpha 
\end{equation}
and the associated reliability state is updated as
\begin{equation}
C_{t+1}(\mathbf{x})=
\frac{C_t(\mathbf{x})^2+w_t(\mathbf{x})^2}
{C_t(\mathbf{x})+w_t(\mathbf{x})},
\quad \forall \mathbf{x}\in m_\alpha 
\end{equation}
where $C_t(x)$ represents the fused reliability associated with the current preference estimate. The update favors evidence with larger reliability weights while retaining historical information from previous viewpoints.

Through repeated updates across navigation steps and viewpoints, $V_t(x)$ forms a multi-view semantic-spatial memory that preserves consistent evidence while reducing the influence of isolated low-reliability observations. As illustrated in Fig.~\ref{fig:preference_map}(C), this reliability-aware updating prevents distant semantic evidence from dominating the global preference map and produces more stable exploration preferences. After each update, the direction-level preference is obtained by averaging $V_{t+1}(\mathbf{x})$ over the corresponding directional mask, and the resulting scores are passed to the subsequent navigation-planning module.

\subsection{Exploration Task Planning Module}

Based on the fused global preference map, CSG-Nav performs hierarchical navigation planning in two stages: direction selection and executable action selection. This module does not introduce an additional fusion mechanism; instead, it converts the fused semantic-spatial preferences from Sec.~\ref{sec:spatial-conf-fusion} into navigation actions. The agent first selects a promising observation direction and generates a corresponding subtask, and then chooses an executable local action within the reachable region.

\subsubsection{Direction Selection}
After the spatial-confidence fusion update, the preference of each candidate direction is obtained by averaging the updated preference map $V_{t+1}(\mathbf{x})$ over the corresponding directional mask. We denote the resulting fused directional preference as $\overline{S}_t(\alpha)$. If the target is detected in one of the candidate views, the corresponding direction is prioritized for the subsequent target-approach phase. Otherwise, the next exploration direction is selected according to

\begin{equation}
\alpha_t^{*} = \arg\max_{\alpha} \overline{S}_t(\alpha)
\end{equation}
The candidate angles are defined in the horizontal plane relative to the current agent heading, with $0^\circ$ aligned with the forward direction and angles increasing clockwise. After selecting $\alpha_t^*$, the agent performs the minimum required rotation to align with the selected view.

Once aligned, the task planner generates a natural-language subtask instruction $S_t^{sub}$. The planner takes as input the selected observation $I_{\alpha_t^*}$, the previous subtask $S_{t-1}^{sub}$, the corresponding semantic rationale $R_t(\alpha_t^*)$, and the target category $T$. It outputs the current subtask, such as "go to the hallway" or "approach the sofa", together with a flag $G_{\mathrm{flag}}$. The flag indicates whether the target is considered visually detected and whether the agent should switch from regular exploration to the target-approach and verification phase.

\subsubsection{Action Selection}

We adopt the candidate-action proposal strategy of WMNav~\citep{11246684} as the execution layer of CSG-Nav. Given the selected direction and subtask, the agent estimates the locally accessible region from RGB-D observations and samples candidate movement directions $\theta_i$ within the current field of view. For each direction, the reachable travel distance $r_i$ is estimated from the depth observation and accessibility mask, forming a polar candidate action $a_i = (r_i,\theta_i)$. 

The candidate set is then filtered to remove actions with insufficient travel distance or collision risk, while spatially close candidates are merged to reduce redundancy. The remaining actions form the executable set $A_t^{final}$. When $G_{\mathrm{flag}}$ is true, indicating that the target has been detected, the spacing constraint is relaxed to generate a denser candidate set for more precise local adjustment.

To make the spatial meaning of each candidate directly observable to the VLM, the actions in $A_t^{final}$ are projected onto the current RGB image and visualized using numbered arrows, producing the candidate-action image $I_t^{act}$. The numbered candidates therefore correspond directly to the actions in $A_t^{final}$. Given $I_t^{act}$, the current subtask $S_t^{sub}$, and the target $T$, the VLM selects an action index $k$, and the agent executes the corresponding action $a_k=(r_k,\theta_k)$.

\subsection{Semantic-Geometric Decision Fusion Module}

\begin{figure*}[ht]
\centering
\includegraphics[width=0.9\textwidth]{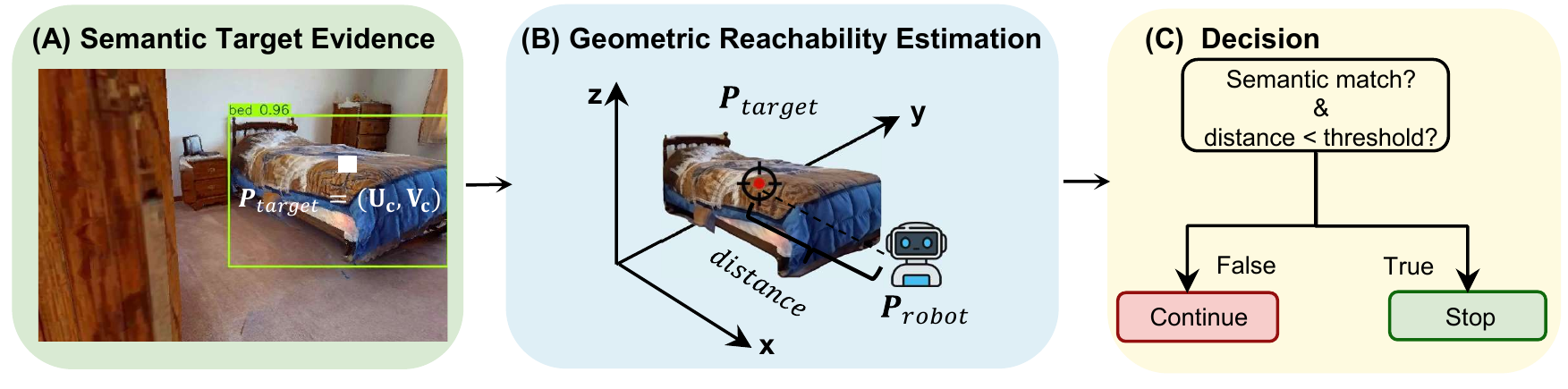}
\caption{Semantic-geometric decision fusion. CSG-Nav combines object-level semantic detection with depth-based 3D localization to evaluate whether the target satisfies the navigation arrival condition. The agent stops only when the detected target matches the goal category and satisfies the geometric distance threshold.}
\label{fig:arrival_verification}
\end{figure*}

Semantic recognition alone is insufficient for reliable goal termination because detecting the target in the current observation does not necessarily indicate that the agent has reached it according to the navigation success criterion. CSG-Nav therefore performs semantic-geometric decision fusion before issuing the stop decision. As illustrated in Fig.~\ref{fig:arrival_verification}, the process consists of semantic target detection, depth-based 3D localization, and final stop-decision evaluation.

When the planning module indicates that the target has been visually detected, we select the highest-confidence target-matching detection with a confidence score of at least 0.25. Its bounding-box center $(u,v)$ and associated depth are used to estimate the target's 3D position, which is then transformed into the global coordinate system using the current agent pose, as illustrated in Fig.~\ref{fig:arrival_verification}(B).

To reduce the effect of detection fluctuation and multi-target ambiguity, the agent maintains a target location buffer $\mathcal{B}$ to record the estimated target positions in the last three frames. The average target position is used to compute the Euclidean distance $D_{dist}$ between the agent and the target. The stop decision is then given by:
\begin{equation}
\text{StopFlag} =
\begin{cases}
\text{True},  & \text{if } P_{sem}=1 \text{ and } D_{\text{dist}} < D_{\text{thres}}, \\
\text{False}, & \text{otherwise}.
\end{cases}
\end{equation}
Here, \(P_{\mathrm{sem}}=1\) indicates that the current detection matches the target category, and \(D_{\mathrm{thres}}=1.0\) m is the Euclidean-distance threshold used for the stopping decision.

If the target is detected but the geometric arrival condition is not yet satisfied, the agent remains in the target-approach phase and continues local adjustment and re-observation. In this way, semantic target identity and 3D proximity are jointly considered before termination, reducing premature stopping caused by semantic-only decisions.

\begin{algorithm}[t]
\small
\caption{CSG-Nav Inference Loop for Zero-Shot Object Navigation}
\label{alg:CSG-Nav}
\textbf{Input:} target object $T$ \\
\textbf{Output:} navigation action $(r_t,\theta_t)$ or \textbf{STOP}

\begin{algorithmic}[1]

\State Initialize preference map $V_0$, confidence map $C_0$,
previous subtask $S_{\mathrm{prev}}^{sub}\gets\emptyset$

\While{episode not terminated}

    \State $\{I_\alpha\}
    \gets \textsc{PanoramicObserve}()$

    \State $\{S_t(\alpha),R_t(\alpha)\}
    \gets \textsc{SemanticSourceFusion}(\{I_\alpha\},T)$

    \State $(V_{t+1},C_{t+1},\{\overline{S}_t(\alpha)\})
    \gets
    \textsc{SpatialConfidenceFusion}
    (\{S_t(\alpha)\},V_t,C_t)$

    \If{$\exists\,\alpha:S_t(\alpha)=10$}
        \State $\alpha_t^*
        \gets \textsc{TargetAngleSelect}(\{S_t(\alpha)\})$
    \Else
        \State $\alpha_t^*
        \gets \arg\max_{\alpha}\overline{S}_t(\alpha)$
    \EndIf

    \State \textsc{AlignTo}$(\alpha_t^*)$

    \State $(S_t^{sub},G_{flag})
    \gets
    \textsc{SubtaskPlanning}
    (I_{\alpha_t^*},S_{\mathrm{prev}}^{sub},
    R_t(\alpha_t^*),T)$

    \If{$G_{flag}=\textsc{true}$}
        \State $\text{StopFlag}
        \gets
        \textsc{SemanticGeometricFusion}
        (I_{\alpha_t^*},T)$

        \If{$\text{StopFlag}=\textsc{true}$}
            \State \textbf{return STOP }
        \EndIf
    \EndIf

    \State $A_t^{final}
    \gets
    \textsc{ActionProposal}
    (I_{\alpha_t^*},S_t^{sub},G_{flag})$

    \State $(r_t,\theta_t)
    \gets
    \textsc{ActionDecision}
    (I_{\alpha_t^*},A_t^{final},S_t^{sub},T)$

    \State \textsc{Execute}$(r_t,\theta_t)$

    \State $S_{\mathrm{prev}}^{sub}\gets S_t^{sub}$

\EndWhile

\end{algorithmic}
\end{algorithm}

The complete inference procedure of CSG-Nav is summarized in Algorithm~\ref{alg:CSG-Nav}. At each step, semantic-source fusion produces directional semantic scores, which are integrated with distance-conditioned reliability and accessibility constraints to update the global preference map. The planning module then selects a direction, generates a subtask, and chooses an executable action. Once the target is detected, semantic-geometric decision fusion determines whether the stopping condition is satisfied; otherwise, the agent continues exploration or target approach.

\section{Simulated Experiments}
\label{sec:simulated_experiments}

\subsection{Experimental Setup}

\subsubsection{Datasets}

We evaluate CSG-Nav using the Habitat~\citep{szot2021habitat} simulator on three standard object-goal navigation benchmarks: Habitat-Matterport 3D (HM3D)~\citep{ramakrishnan2habitat}, HM3Dv2~\citep{yadav2023habitat}, and Matterport 3D (MP3D)~\citep{Matterport3D}. HM3Dv1 provides 2000 validation episodes from 20 scenes and covers 6 object categories. HM3Dv2 improves scene quality and semantic annotation scale, providing 1000 validation episodes from 36 scenes and 6 object categories. MP3D contains 2195 episodes from 11 scenes and includes 21 object categories. 

These datasets cover different scene distributions, annotation settings, and target-category scales, enabling evaluation across diverse ObjectNav benchmarks.

\subsubsection{Metrics}

We report Success Rate (SR) and Success weighted by Path Length (SPL). SR measures the percentage of episodes in which the agent successfully stops near the target object. SPL further evaluates navigation efficiency by comparing the executed path length with the shortest path length. Therefore, SR reflects the reliability of goal reaching, while SPL reflects both success and path efficiency. Since CSG-Nav emphasizes physically grounded stopping, SR is particularly important for evaluating whether semantic-geometric fusion reduces false arrival decisions.

\subsubsection{Baselines}

We compare CSG-Nav with representative object-goal navigation methods on HM3Dv1~\citep{ramakrishnan2habitat}, HM3Dv2~\citep{yadav2023habitat}, and MP3D~\citep{Matterport3D}. The baselines cover both zero-shot and training-based settings.

For zero-shot object-goal navigation, we include perception- and semantic-map-based methods, including ZSON~\citep{majumdar2022zson}, CoW~\citep{gadre2023cows}, VLFM~\citep{yokoyama2024vlfm}, PSL~\citep{sun2024prioritized}, PixNav~\citep{cai2024bridging} and OpenFMNav~\citep{kuang2024openfmnav}. We also compare with reasoning- and foundation-model-based approaches, including ESC~\citep{zhou2023esc}, L3MVN~\citep{Yu_2023}, EIK-Nav~\citep{jiang2025eik}, UniGoal~\citep{yin2025unigoal}, ReMemNav~\citep{wu2026rememnav} and WMNav~\citep{11246684}, which further incorporates world-model-based prediction into navigation planning.

For training-based comparison, we include Habitat-Web~\citep{ramrakhya2022habitat}, PIRLNav~\citep{ramrakhya2023pirlnav}, SemEXP~\citep{chaplot2020object}, PONI~\citep{ramakrishnan2022poni}, and ProcTHOR~\citep{deitke2022procthor}. Together, these baselines provide comparisons across perception-driven, semantic-map, reasoning-based, foundation-model, world-model, and supervised policy-learning paradigms.

\subsubsection{Implementation Details}

We set the maximum number of navigation steps to 40. The agent moves with a minimum step length of 0.5 m and a maximum movement distance of 1.7 m. The agent height is set to 0.88 m, and the agent radius is set to 0.18 m. The RGB-D camera is mounted with a downward tilt of approximately 14 degrees. It has a field of view of 79 degrees and outputs images at a resolution of $480 \times 640$. The navigation success threshold is set to $S_{thres}=1.0$ m. An episode is considered successful if the shortest path distance between the agent and the target object is less than 1.0 m when the agent stops.

For confidence-aware preference map generation, we initialize the preference value as $V_0(\mathbf{x})=10$ and the confidence value as $C_0(\mathbf{x})=0.2$. The former provides an optimistic exploration prior for unobserved regions, while the latter allows subsequent observations to have greater influence on preference-map updates. Semantic observations are then fused using distance-conditioned reliability weights.

For the model configuration, CSG-Nav uses Qwen3-VL-4B as the main VLM for scene understanding, subtask decomposition, and action selection, and Qwen3-4B as the LLM for structured reasoning and direction-level semantic scoring. YOLO-World provides explicit object-level detections, detection confidence, and bounding boxes for semantic-source fusion and semantic-geometric decision fusion. The VLM and LLM are invoked sequentially according to their respective roles in the navigation pipeline. We adopt this configuration as the default compact model setting for CSG-Nav. Its computational efficiency is further evaluated in Sec.~\ref{sec:efficiency} through navigation-step latency, model inference latency, spatial-fusion overhead, and peak GPU memory, with an additional comparison against a larger configuration using Qwen2.5-VL-7B as the VLM and Qwen3-8B as the LLM.

\subsection{Overall Results}

CSG-Nav achieves competitive performance in HM3Dv1, HM3Dv2 and MP3D, as summarized in Table~\ref{tab:experiment_result}. It achieves an SR of 58.9\% and 71.6\% on HM3Dv1 and HM3Dv2, outperforming the strongest world-model-based baseline, WMNav, by 0.8 and 1.8 percentage points, respectively. On MP3D, CSG-Nav remains competitive and achieves the second-best SR among zero-shot methods. These results indicate that the proposed confidence-aware multimodal fusion framework demonstrates competitive goal-reaching performance across different benchmark settings.

\begin{table}[ht]
\centering
{\fontsize{10pt}{10pt}\selectfont
\setlength{\tabcolsep}{4.5pt}
\caption{Object Goal Navigation results on HM3Dv1, HM3Dv2 and MP3D. All values are reported as percentages (\%). ZS denotes zero-shot. MP denotes the largest individual-model parameter count when publicly available; for proprietary models with undisclosed parameter counts, the corresponding model names are reported instead.}
\label{tab:experiment_result}
\begin{tabular}{lcccccccc}
\toprule
\multirow{2}{*}{Method} & \multirow{2}{*}{ZS} & \multirow{2}{*}{MP} & \multicolumn{2}{c}{HM3Dv1} & \multicolumn{2}{c}{HM3Dv2} & \multicolumn{2}{c}{MP3D} \\
\cmidrule(lr){4-5} \cmidrule(lr){6-7} \cmidrule(lr){8-9}
& & & SR$\uparrow$ & SPL$\uparrow$ & SR$\uparrow$ & SPL$\uparrow$ & SR$\uparrow$ & SPL$\uparrow$ \\
\midrule

Habitat-Web\citep{ramrakhya2022habitat} & $\times$ &  & 41.5 & 16.0 & - & - & 31.6 & 8.5 \\
ProcThor\citep{deitke2022procthor} & $\times$ &  & 54.4 & 31.8 & - & - & - & - \\
PONI\citep{ramakrishnan2022poni} & $\times$ &  & - & - & - & - & 31.8 & 12.1 \\
SemEXP\citep{chaplot2020object} & $\times$ &  & - & - & - & - & 36.0 & 14.4 \\
PIRLNav\citep{ramrakhya2023pirlnav} & $\times$ &  & 64.1 & 27.1 & 66.0 & 27.0 & - & - \\

\midrule

ZSON\citep{majumdar2022zson} & $\checkmark$ & $< 1\,\text{B}$ & 25.5 & 12.6 & - & - & 15.3 & 4.8 \\
PSL\citep{sun2024prioritized} & $\checkmark$ & $< 1\,\text{B}$ & 42.4 & 19.2 & - & - & 18.9 & 6.4 \\
PixNav\citep{cai2024bridging} & $\checkmark$ & GPT-4 & 37.9 & 20.5 & - & - & - & - \\
VLFM\citep{yokoyama2024vlfm} & $\checkmark$ & $> 1\,\text{B}$ & 52.5 & 30.4 & 63.6 & 32.5 & 36.2 & 15.9 \\
CoW\citep{gadre2023cows} & $\checkmark$ & $< 1\,\text{B}$ & - & - & - & - & 9.2 & 4.9 \\
ESC\citep{zhou2023esc} & $\checkmark$ & $< 1\,\text{B}$ & 39.2 & 22.3 & - & - & 28.7 & 14.2 \\
L3MVN\citep{Yu_2023} & $\checkmark$ & $< 1\,\text{B}$ & 50.4 & 23.1 & 36.3 & 15.7 & - & - \\
OpenFMNav\citep{kuang2024openfmnav} & $\checkmark$ & GPT-4/4V & 54.9 & 24.4 & - & - & 37.2 & 15.7 \\
SG-Nav-GPT\citep{yin2024sg} & $\checkmark$ & GPT-4 & 54.0 & 24.9 & 49.6 & 25.5 & 40.2 & 16.0 \\
VoroNav\citep{wu2024voronavvoronoibasedzeroshotobject} & $\checkmark$ & GPT-3.5-turbo & 42.0 & 26.0 & - & - & - & - \\
TopV-Nav\citep{zhong2025topvnavunlockingtopviewspatial} & $\checkmark$ & $7\,\text{B}$ & 45.9 & 28.0 & - & - & 31.9 & 16.1 \\
EIK-Nav\citep{jiang2025eik} & $\checkmark$ & $72\,\text{B}$ & 54.8 & 31.5 & - & - & 37.1 & 17.4 \\
UniGoal\citep{yin2025unigoal} & $\checkmark$ & $7\,\text{B}$ & 54.5 & 25.1 & - & - & 41.0 & 16.4 \\
ReMemNav \citep{wu2026rememnav} & $\checkmark$ & $4\,\text{B}$ & - & - & 56.2 & 21.8 & - & - \\
WMNav\citep{11246684} & $\checkmark$ & Gemini 1.5 Pro & \underline{58.1} & 31.2 & \underline{69.8} & 27.8 & \textbf{45.4} & 17.2 \\

\midrule
Ours & $\checkmark$ & $4\,\text{B}$ & \textbf{58.9} & 24.5 & \textbf{71.6} & 29.5 & \underline{44.3} & 17.0 \\
\bottomrule
\end{tabular}
}
\end{table}

A notable advantage of CSG-Nav is its compact model configuration. Many existing LLM/VLM-based methods rely on large closed-source models, such as GPT or Gemini-scale models. In contrast, CSG-Nav achieves competitive zero-shot performance using 4B-scale VLM and LLM components. This suggests that explicitly fusing semantic evidence, confidence-weighted spatial memory, and geometric verification can partially reduce the dependence on extremely large foundation models.

Among all compared zero-shot methods, CSG-Nav achieves the highest reported SR among the compared zero-shot methods on HM3Dv1 and HM3Dv2. However, its SPL is lower than some baselines. We attribute this mainly to the conservative nature of our semantic-geometric stopping strategy. Unlike methods that may stop once strong semantic target cues are observed, CSG-Nav requires additional geometric verification before termination. When the target is detected but the geometric stopping condition is not yet satisfied, the agent may perform local adjustment and re-observation to confirm arrival. This behavior improves stopping reliability and contributes to higher SR, but it also increases redundant motion near the goal and reduces SPL. In addition, the compact 4B LLM may limit the quality of long-horizon reasoning, which can further affect planning efficiency.

Compared with training-based methods, PIRLNav achieves the highest SR on HM3Dv1 by using 77k human demonstrations. Nevertheless, CSG-Nav narrows the gap to about 5 percentage points without task-specific training. This demonstrates the effectiveness of the proposed zero-shot fusion framework under limited supervision.

\subsection{Ablation Study}

To evaluate the contribution of each fusion component, we conduct controlled ablation studies on HM3Dv2. We first compare different spatial-fusion strategies while using VLM-based semantic scoring and semantic-only stopping. We then fix the Gaussian spatial-fusion strategy and evaluate different combinations of semantic sources, with the stopping strategy kept unchanged. Finally, we retain the Gaussian spatial fusion and the full semantic-source configuration and compare different arrival-verification strategies. This sequential design enables staged comparisons of different fusion configurations while keeping the remaining navigation settings fixed within each comparison.

All ablation experiments are conducted on a fixed subset of 100 HM3Dv2 episodes. We partition the ordered validation set into 50 subsets and select the first two episodes from each subset. The selected episodes cover all 36 validation scenes and all six target-object categories. The same subset is used for the spatial-confidence, semantic-source, and semantic-geometric fusion ablations, as well as the Gaussian-parameter sensitivity and failure analyses. Each configuration is evaluated once, so the reported results are single-run measurements. Accordingly, the ablation results are presented as controlled comparisons across different fusion configurations.

\paragraph{\textbf{Spatial-Confidence Fusion Ablation}}

We first evaluate the contribution of spatial-confidence fusion while keeping the semantic-scoring and stopping strategies fixed. The purpose of this ablation is to examine whether distance-conditioned reliability improves the aggregation of semantic evidence into the global preference map. In particular, long-range observations may provide less reliable semantic evidence than nearby observations, so assigning equal influence to all projected observations can introduce unstable preferences during map updating.

As shown in Table~\ref{tab:ablation_of_confidence}, the baseline follows the WMNav-style value-map construction. Uniform weighting assigns the same confidence weight to the projected semantic evidence, while linear and Gaussian decay progressively reduce the influence of observations with increasing distance. On the evaluated subset, Gaussian distance weighting achieves a higher SR than the baseline, whereas uniform weighting and linear decay do not show the same pattern. These results suggest that the choice of distance-conditioned reliability weighting affects navigation outcomes.

\begin{table}[h]
\centering
\caption{Effects of different spatial-fusion strategies on HM3Dv2.}
\label{tab:ablation_of_confidence}
\begin{tabular}{lcc}
\toprule
Spatial fusion strategy & SR$\uparrow$ & SPL$\uparrow$ \\
\midrule
Baseline          & 69 & 28.19 \\
Uniform weighting & 67 & 26.38 \\
Linear decay      & 64 & 27.64 \\
Gaussian decay    & \textbf{71} & 28.06 \\
\bottomrule
\end{tabular}
\end{table}

\paragraph{\textbf{Semantic-Source Fusion Ablation}}

We next examine different semantic scoring configurations while keeping the Gaussian spatial-confidence fusion and semantic-only stopping strategies fixed.
Table~\ref{tab:ablation_of_llm} summarizes
the semantic inputs and scoring models of the four
configurations. The first and third configurations
use the VLM for direct RGB-based scoring, with
the latter additionally incorporating YOLO-World
detections to assess the effect of object-level
evidence. The second configuration uses only
YOLO-World detections as textual inputs to the
LLM. The fourth configuration further integrates
scene-level descriptions extracted by the VLM
with YOLO-World detections, allowing the LLM
to jointly reason over both semantic sources
and assign direction-level scores.

\begin{table}[h]
\centering
\caption{Effects of different semantic-source fusion configurations on navigation performance in HM3Dv2.}
\label{tab:ablation_of_llm}
\begin{tabular}{lccc}
\toprule
Semantic Source & Scoring & SR$\uparrow$ & SPL$\uparrow$ \\
\midrule
Raw RGB & VLM & 71 & 28.06 \\
YOLO-World detections & LLM & 62 & 20.57 \\
Raw RGB + YOLO-World detections & VLM & 72 & 29.17 \\
VLM scene descriptions + YOLO-World detections & LLM & \textbf{75} & 27.62 \\
\bottomrule
\end{tabular}
\end{table}

As shown in Table~\ref{tab:ablation_of_llm}, integrating VLM-generated scene-level descriptions
with YOLO-World object-level detections through
LLM-assisted scoring achieves the highest SR
on the evaluated subset. This result suggests
that combining complementary scene-level and
object-level semantic evidence can benefit
navigation performance.

\paragraph{\textbf{Semantic-Geometric Fusion Ablation}}

We finally evaluate the contribution of semantic-geometric arrival verification while retaining the Gaussian spatial-confidence fusion and the full semantic-source configuration from the preceding ablations. The purpose of this ablation is to examine whether combining semantic target recognition with geometric localization improves goal termination compared with semantic-only stopping.

We compare three arrival-verification strategies: semantic-only verification, VLM-based semantic-geometric verification using Scaffolding~\citep{lei2025scaffolding}, and the proposed YOLO-based semantic-geometric verification. As shown in Table~\ref{tab:ablation_of_arrival_verification}, the semantic-only strategy stops based on semantic recognition alone, while the two semantic-geometric variants additionally estimate the target location from the image and depth observations. The Scaffolding-based strategy relies on VLM-based target-center estimation which may suffer from hallucination or ambiguous visual grounding, whereas our approach uses explicit YOLO-World detection combined with depth-based 3D localization. The proposed YOLO-based semantic-geometric verification improves navigation performance compared with semantic-only stopping. This suggests that incorporating explicit object localization and depth-based geometric constraints contributes to more reliable goal-termination decisions.

Overall, the three ablation studies highlight the distinct roles of the proposed fusion mechanisms within CSG-Nav. Spatial-confidence fusion incorporates distance-conditioned reliability into preference-map construction, semantic-source fusion combines complementary semantic evidence through LLM-assisted scoring, and semantic-geometric verification integrates object detection with 3D localization for goal termination. Together, these mechanisms enable reliability-aware information fusion across successive stages of zero-shot object-goal navigation.

\begin{table}[htbp]
\centering
\caption{Effects of semantic-geometric decision fusion strategies on the fixed HM3Dv2 ablation subset.}
\label{tab:ablation_of_arrival_verification}
\begin{tabular}{lcc}
\toprule
Arrival verification strategy & SR$\uparrow$ & SPL$\uparrow$ \\
\midrule
Semantic only                         & 75 & 27.62 \\
VLM-based semantic-geometric~\citep{lei2025scaffolding}
                                      & 74 & 27.30 \\
YOLO-based semantic-geometric (ours)  & \textbf{77} & \textbf{28.99} \\
\bottomrule
\end{tabular}
\end{table}
\FloatBarrier

\subsection{Reliability and Sensitivity Analysis}
\label{sec:reliability and sensitivity analysis}
\paragraph{\textbf{Distance-dependent reliability}}

We examine whether perception reliability degrades with observation distance, motivating the distance-conditioned weighting used in spatial-confidence fusion. Target perception is evaluated from 1 to 5~m for a small unobstructed plant and a large chair partially occluded by a doorway. At each distance, five nearby viewpoints are sampled using the same perception pipeline as in navigation, and we report YOLO-World detection rate and confidence together with VLM recognition rate.
As shown in Table~\ref{tab:distance_reliability}, perception reliability generally decreases at longer distances, particularly beyond 3--4~m. These results support using distance as a practical reliability cue for spatial evidence fusion, without assuming that perception uncertainty follows an exact Gaussian distribution.
\begin{table}[htbp]
\centering
\caption{Target perception reliability at different observation distances. Det. denotes detection rate, Conf. denotes average detection confidence, and Recog. denotes recognition rate.}
\label{tab:distance_reliability}
\begin{tabular}{llccc}
\toprule
Target / Setting & Distance & \makecell{YOLO\\Det.} & \makecell{YOLO\\Conf.} & \makecell{VLM\\Recog.} \\
\midrule
\multirow{5}{*}{\makecell{Small plant,\\unobstructed}}
& 1~m & 5/5 & 0.739 & 5/5 \\
& 2~m & 5/5 & 0.780 & 3/5 \\
& 3~m & 5/5 & 0.727 & 3/5 \\
& 4~m & 5/5 & 0.660 & 3/5 \\
& 5~m & 5/5 & 0.563 & 1/5 \\
\midrule
\multirow{5}{*}{\makecell{Large chair,\\doorway occlusion}}
& 1~m & 5/5 & 0.868 & 5/5 \\
& 2~m & 5/5 & 0.892 & 5/5 \\
& 3~m & 5/5 & 0.890 & 5/5 \\
& 4~m & 4/5 & 0.775 & 2/5 \\
& 5~m & 0/5 & -- & 0/5 \\
\bottomrule
\end{tabular}
\end{table}

\paragraph{\textbf{Sensitivity to Gaussian decay parameter $\sigma$}}

\begin{figure}[!htbp]
\centering
\begin{tikzpicture}
\begin{axis}[
    width=0.72\columnwidth,
    height=0.36\columnwidth,
    xlabel={$\sigma$ (m)},
    ylabel={SR (\%)},
    xmin=0.4, xmax=3.1,
    ymin=60, ymax=80,
    xtick={0.5,1.0,1.5,2.0,3.0},
    ytick={60,65,70,75,80},
    grid=major,
    mark size=2.5pt,
]
\addplot[
    mark=*,
    thick
] coordinates {
    (0.5,68)
    (1.0,77)
    (1.5,74)
    (2.0,75)
    (3.0,75)
};
\end{axis}
\end{tikzpicture}
\caption{Sensitivity of navigation performance to the Gaussian decay parameter
$\sigma$ on HM3Dv2.}
\label{fig:sigma_sensitivity}
\end{figure}
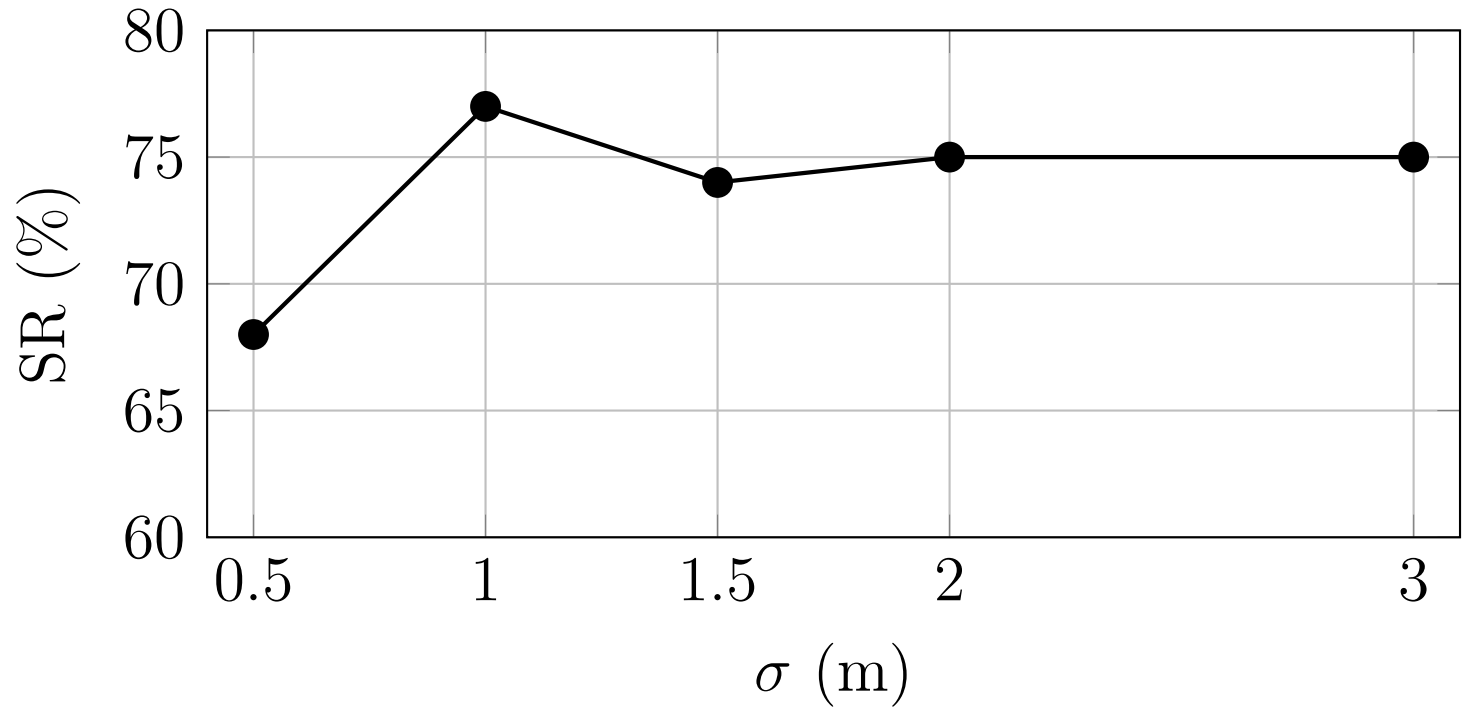

We further evaluate the sensitivity to the Gaussian decay parameter $\sigma$ on the same fixed 100-episode HM3Dv2 subset used in the ablation studies. All other components are kept fixed, and only $\sigma$ is varied. As shown in Fig.~\ref{fig:sigma_sensitivity}, $\sigma=0.5$~m leads to lower SR, whereas performance remains relatively stable for $\sigma$ between 1.0 and 3.0~m. This indicates that CSG-Nav is not highly sensitive to moderate variations in the Gaussian decay scale.

\FloatBarrier

\subsection{Computational Efficiency}
\label{sec:efficiency}
To evaluate the computational cost of CSG-Nav, we compare the default compact configuration with a larger model configuration while keeping the navigation pipeline unchanged. The default configuration uses Qwen3-VL-4B as the VLM and Qwen3-4B as the LLM, whereas the larger configuration uses Qwen2.5-VL-7B and Qwen3-8B, respectively. We report the aggregated VLM/LLM inference latency, YOLO-World inference latency, spatial-confidence fusion latency, semantic-geometric decision-fusion latency, navigability updating and planning latency, and peak GPU memory consumption. Since the VLM and LLM are invoked at different stages of the pipeline, their inference latency is aggregated and reported as VLM/LLM latency. The reported latency and memory statistics are averaged over the 100 fixed navigation episodes for each model configuration.

\begin{table}[t]
\centering
\caption{Computational efficiency of different model configurations. 
All latency values are reported in seconds (s), and peak GPU memory is reported in GB. 
Geo. Fusion denotes semantic-geometric decision fusion, and Nav. updating denotes navigability updating and planning.}
\label{tab:efficiency}
\resizebox{\linewidth}{!}{
\begin{tabular}{lcccccc}
\hline
Configuration
& \makecell{VLM/LLM \\ latency}
& \makecell{YOLO\\latency}
& \makecell{Spatial \\ Fusion}
& \makecell{Geo. \\ Fusion}
& \makecell{Nav. \\ updating}
& \makecell{Peak \\ Memory} \\
\hline
\makecell{Qwen3-VL-4B \\ + Qwen3-4B}
& 1.64
& 0.35
& 0.82
& 0.02
& 16.61
& 7.74 \\

\makecell{Qwen2.5-VL-7B \\ + Qwen3-8B}
& 2.00
& 0.36
& 0.84
& 0.02
& 16.62
& 16.60 \\
\hline
\end{tabular}
}
\end{table}

As shown in Table~\ref{tab:efficiency}, increasing the model scale mainly affects VLM/LLM inference latency and peak GPU memory consumption, while the YOLO-World, spatial-confidence fusion, and semantic-geometric decision-fusion latencies remain relatively stable across the two configurations. In contrast, the largest portion of the per-step computation is associated with navigability updating and planning, which mainly involves system-level map processing and navigation operations. These results indicate that the proposed fusion components introduce limited additional computational overhead, while the compact 4B-scale configuration reduces foundation-model memory usage and inference cost.

\subsection{Failure Analysis}
\label{sec:failure_analysis}

\begin{figure*}[ht]
\centering
\includegraphics[width=0.7\textwidth]{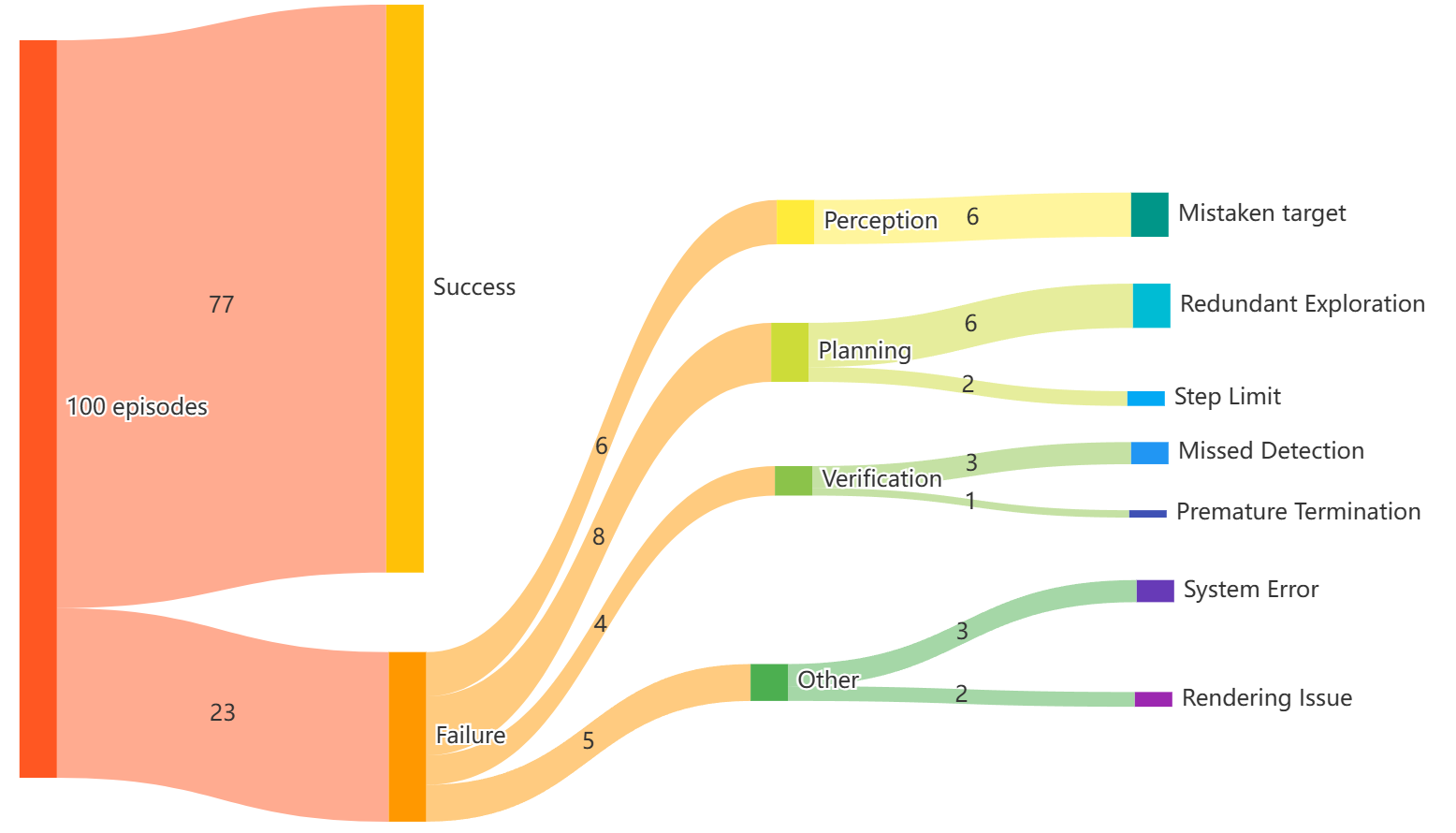}
\caption{Failure analysis of the final CSG-Nav configuration on the fixed HM3Dv2 subset.}
\label{fig:failure_analysis}
\end{figure*}

We further analyze the failure cases of the final CSG-Nav configuration on the same fixed 100-episode HM3Dv2 subset used in the ablation studies. As illustrated in Fig.~\ref{fig:failure_analysis}, unsuccessful episodes are categorized according to their primary cause, including planning, perception, verification, and environment-related failures. Each failed episode is assigned to one category based on its navigation trajectory and execution logs.

Planning-related cases account for the largest proportion of the remaining failures. These cases mainly occur after the global preference map has selected a promising exploration direction, but local navigation becomes difficult in narrow or cluttered regions, or the target area cannot be reached within the maximum step limit in large scenes. This suggests that the remaining planning challenges are mainly associated with local execution and long-horizon navigation rather than semantic preference estimation.

Perception-related cases mainly arise from ambiguous visual evidence. Visually or semantically similar objects may occasionally be confused, such as sofas with beds or benches with chairs. These cases indicate that difficult visual ambiguity remains challenging even after combining object-level detection, scene-level understanding, and relational reasoning.

Verification-related cases are mainly observed during target approach under occlusion or unfavorable viewpoints. In such situations, stable object detection and depth-based localization may be temporarily unavailable, delaying reliable arrival verification. When consistent target observations are available, semantic-geometric fusion provides explicit geometric evidence for the stopping decision.

Finally, some unsuccessful episodes are associated with challenging scene layouts or evaluation conditions, such as difficult-to-exit regions or cases close to the success boundary. Overall, the failure distribution shows that the remaining challenges are concentrated mainly in local navigation, ambiguous observations, and target visibility, providing clear directions for further improving the navigation pipeline.

\subsection{Qualitative Analysis}
\label{sec:qualitative_analysis}

\begin{figure*}[ht]
\centering
\includegraphics[width=0.7\textwidth]{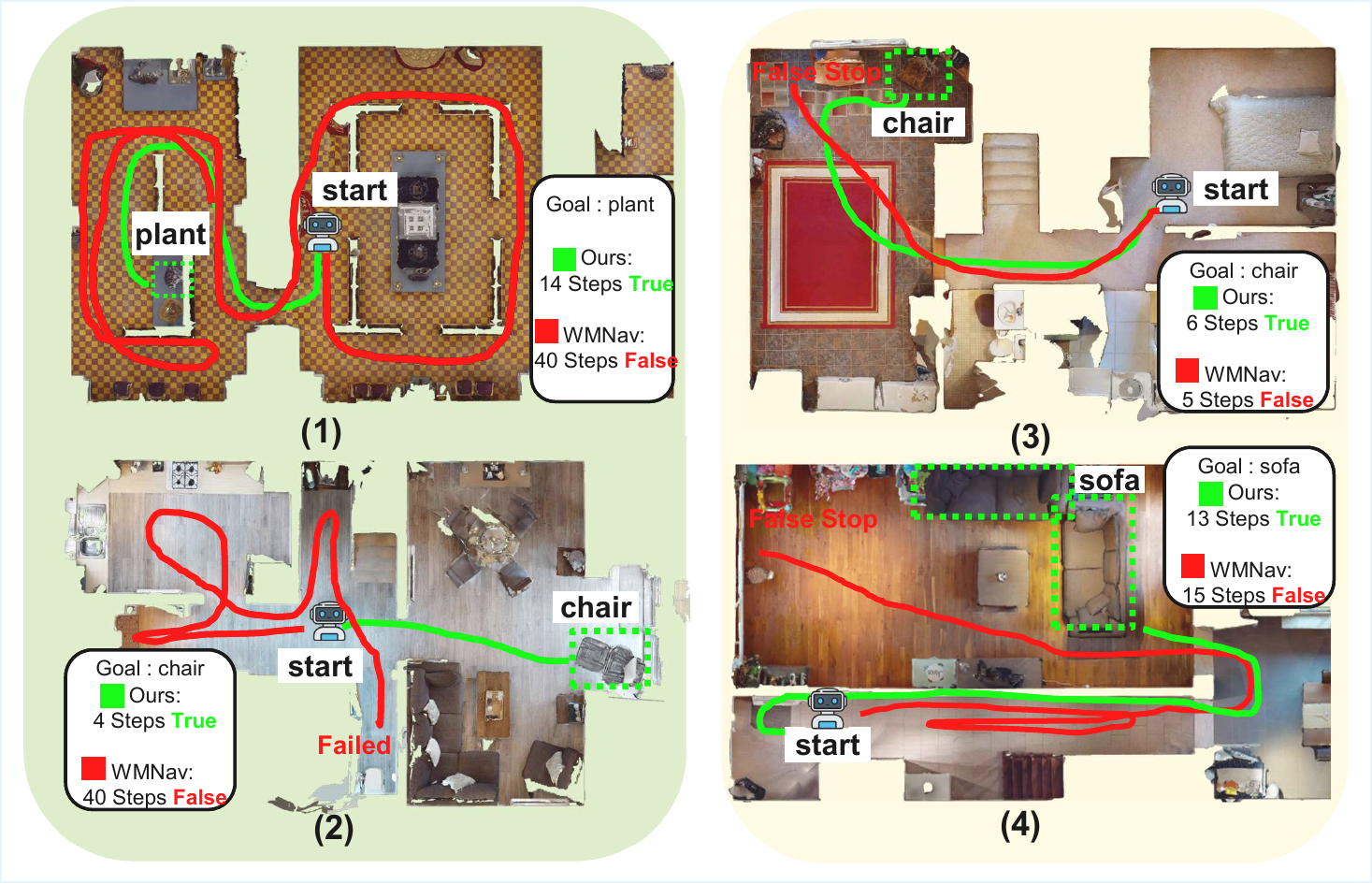}
\caption{Navigation trajectory comparison on HM3Dv2. CSG-Nav is shown in green and WMNav in red.}
\label{fig:compare_path}
\end{figure*}

We further qualitatively examine how the proposed fusion mechanisms affect target perception, exploration, and arrival verification in representative HM3Dv2 scenes. The navigation trajectories of CSG-Nav and WMNav are compared in Fig.~\ref{fig:compare_path}, where the green and red paths denote CSG-Nav and WMNav, respectively.

In case (1), the target plant is small and weakly salient. WMNav passes the target without successfully identifying it, whereas CSG-Nav benefits from complementary semantic evidence: YOLO-World provides explicit object-level detections while the VLM contributes scene-level context. This combination improves target perception when a single semantic source is insufficient.

In case (2), the target is not detected from the initial observation. With spatial-confidence fusion, distant semantic evidence receives lower reliability during preference-map updating, allowing the direction to remain available for subsequent exploration. As the agent approaches, the target becomes more reliably observed and is successfully identified. In contrast, WMNav moves away from the region and does not locate the target within the maximum step limit. This example illustrates how distance-conditioned reliability weighting helps preserve useful exploration preferences under uncertain long-range observations.

Cases (3) and (4) illustrate semantic-geometric arrival verification. Although both methods obtain semantic evidence for the target, WMNav terminates navigation while the target remains beyond the desired stopping distance. CSG-Nav additionally uses depth-based 3D localization and stops only when the semantic match is accompanied by a satisfied 3D proximity condition.

Overall, these examples qualitatively demonstrate the complementary roles of the three fusion components in CSG-Nav: semantic-source fusion strengthens target perception, spatial-confidence fusion improves exploration preference estimation, and semantic-geometric fusion provides more reliable goal termination.

\section{Real-world Experiments}
\label{sec:real-world_experiments}

\subsection{Experiment Setup}

\begin{figure*}[ht]
\centering
\includegraphics[width=0.9\textwidth]{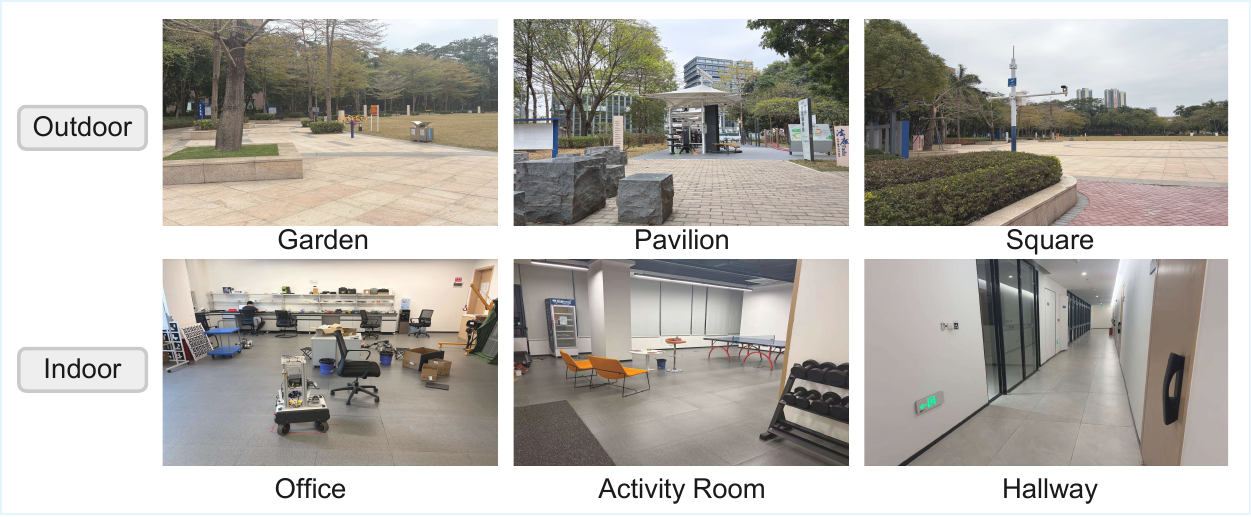}
\caption{Real-world scene demonstrations. The outdoor environment is a park with multiple regions, while the indoor environments include an office, a hallway, and an activity room.}
\label{fig:realworld_scene}
\end{figure*}

We further evaluate CSG-Nav in real indoor and outdoor environments to examine its transferability from simulation to practical robotic navigation. As shown in Fig.~\ref{fig:realworld_scene}, the indoor environments include an office, a hallway, and an activity room, while the outdoor environment is a park with multiple regions. These scenes contain targets with different object scales, viewpoints, lighting conditions, and background complexity.

The robot is equipped with an RGB-D camera, a wheeled mobile base, and an NVIDIA Jetson AGX Orin 64GB computing unit. The same model configuration and navigation pipeline used in simulation, including semantic-source fusion, spatial-confidence fusion, RGB-D-based candidate-action proposal and visualization, and semantic-geometric verification, are retained for real-world deployment. The perception and foundation-model components are executed locally on the robot, allowing CSG-Nav to operate without continuous dependence on external network connectivity.

\subsection{Results}

\begin{table}[htbp]
\centering
\setlength{\tabcolsep}{4pt}
\caption{Real-world navigation results in outdoor and indoor environments.}
\label{tab:real_world_results}
\begin{tabular}{llcccccc}
\toprule
\multirow{2}{*}{Scene} & \multirow{2}{*}{Goal} & \multicolumn{5}{c}{Test} & \multirow{2}{*}{SR} \\
\cmidrule(lr){3-7}
&  & Test 1 & Test 2 & Test 3 & Test 4 & Test 5 &  \\
\midrule
\multirow{5}{*}{Outdoor}
& Fire box             & $\times$     & $\checkmark$ & $\checkmark$ & $\times$     & $\times$     & 2/5 \\
& Trash can            & $\checkmark$ & $\times$     & $\times$     & $\times$     & $\checkmark$ & 2/5 \\
& Wooden bench         & $\times$     & $\times$     & $\times$     & $\checkmark$ & $\times$     & 1/5 \\
& Electrical box       & $\times$     & $\times$     & $\checkmark$ & $\times$     & $\times$     & 1/5 \\
& Stone bench          & $\checkmark$ & $\checkmark$ & $\times$     & $\checkmark$ & $\checkmark$ & 4/5 \\
\midrule
\multirow{5}{*}{Indoor}
& Elevator             & $\times$     & $\checkmark$ & $\checkmark$ & $\times$     & $\checkmark$ & 3/5 \\
& Toilet               & $\times$     & $\times$     & $\times$     & $\checkmark$ & $\times$     & 1/5 \\
& Water cooler         & $\checkmark$ & $\checkmark$ & $\checkmark$ & $\checkmark$ & $\checkmark$ & 5/5 \\
& Ping-pong table      & $\times$     & $\checkmark$ & $\checkmark$ & $\checkmark$ & $\times$     & 3/5 \\
& Exercise equipment   & $\times$     & $\checkmark$ & $\times$     & $\checkmark$ & $\checkmark$ & 3/5 \\
\bottomrule
\end{tabular}
\end{table}

We conduct five trials for each target category, with the results summarized in Table~\ref{tab:real_world_results}. CSG-Nav successfully completes navigation for multiple target types in both indoor and outdoor environments, while performance varies with object appearance, scale, visibility, and scene layout.

In the outdoor park, the evaluated targets include a fire box, trash can, wooden bench, electrical box, and stone bench. The stone bench achieves 4/5 successful trials, benefiting from its larger visual extent and relatively stable observation geometry. Smaller or less distinctive targets are more sensitive to viewpoint and background complexity.

The indoor experiments include a water cooler, elevator, toilet, ping-pong table, and exercise equipment. The water cooler achieves 5/5 successful trials in a compact office environment, while targets such as the elevator and toilet require longer navigation through corridors or multiple regions. These cases place greater demands on accumulated spatial evidence and long-horizon exploration.

We additionally evaluate a high-mounted surveillance camera as a challenging case. Because the target is positioned well above the robot-mounted RGB-D camera and frequently lies outside its effective field of view, successful observation from standard navigation viewpoints is difficult. This case highlights the importance of viewpoint coverage for real-world object-goal navigation.

\begin{figure*}[ht]
\centering
\includegraphics[width=0.99\textwidth]{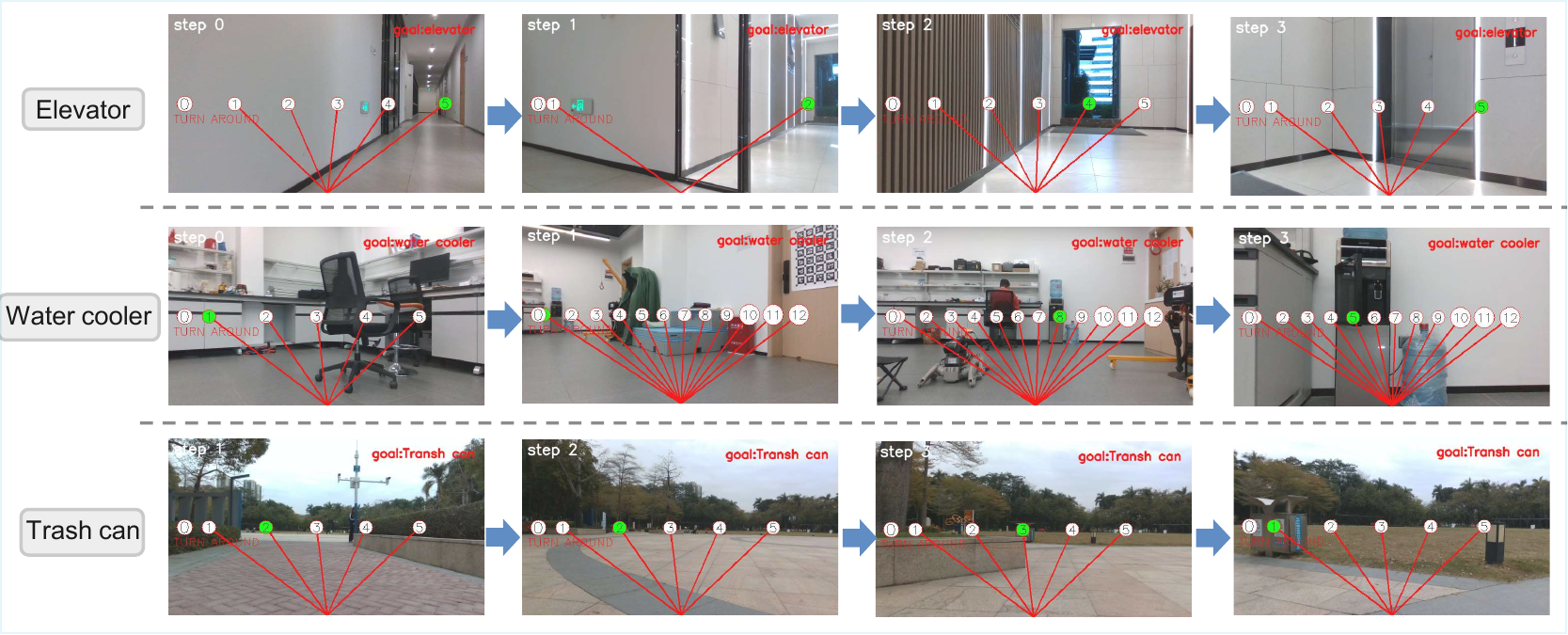}
\caption{Real-world deployment of CSG-Nav for object-goal navigation. Snapshots show the navigation process from start to goal across indoor and outdoor environments, demonstrating the practical feasibility of confidence-aware multimodal fusion in real robotic scenarios.}
\label{fig:navigation_process}
\end{figure*}

Representative navigation processes are shown in Fig.~\ref{fig:navigation_process}. Overall, the experiments demonstrate the feasibility of transferring CSG-Nav from simulation to real indoor and outdoor environments with local onboard execution. The same semantic-source, spatial-confidence, and semantic-geometric fusion pipeline is retained across these scenarios without task-specific retraining.

\section{Discussion and Conclusions}
\label{sec:conclusions}

In this paper, we propose CSG-Nav, a zero-shot object-goal navigation framework based on confidence-aware multimodal information fusion. CSG-Nav integrates heterogeneous semantic evidence, distance-conditioned spatial reliability, and RGB-D geometry through three fusion levels: semantic-source fusion for direction scoring, spatial-confidence fusion for preference-map construction, and semantic-geometric fusion for arrival verification.

Experiments on HM3D and MP3D show competitive zero-shot navigation performance, while component-level ablations illustrate the effects of different fusion configurations on navigation outcomes. Spatial-confidence fusion reduces the influence of less reliable long-range observations, semantic-source fusion benefits from complementary object-level and scene-level evidence, and semantic-geometric fusion improves goal termination through depth-based 3D localization. Reliability and sensitivity analyses further support distance as a practical reliability cue, while real-world experiments demonstrate the feasibility of transferring the same fusion pipeline to indoor and outdoor robotic environments with local onboard execution.

These results suggest that object-level detection, scene-level understanding, and language-based reasoning provide complementary semantic information; reliability-aware spatial aggregation improves exploration preference estimation; and geometric proximity provides an additional signal for reliable arrival verification. Together, these findings support reliability-aware multimodal fusion across multiple stages of zero-shot navigation.

Several challenges remain. The current spatial-confidence module relies on a hand-designed distance-conditioned prior and fixed confidence initialization rather than learned or calibrated uncertainty estimation. Additional sensitivity experiments show that the initial preference $V_0$ substantially influences navigation performance. The initial confidence $C_0$ also affects the results. These findings highlight the sensitivity of the current heuristic initialization and motivate
future work on adaptive confidence estimation
and updating strategies. Perception can also become less stable for high, occluded, or out-of-view targets, which motivates active viewpoint adjustment and cross-time-step target tracking. In addition, the current real-world experiments demonstrate deployment feasibility but do not yet provide a systematic robustness assessment under controlled lighting variation, dynamic environments, or sensor degradation, nor matched baseline and component-level comparisons on the same robotic platform. Future work will therefore include controlled real-world evaluations under these environmental and sensing variations, together with baseline comparisons and component ablations, to assess system robustness and quantify the contribution of each fusion mechanism.

\section*{Acknowledgments}
This work was supported by the Shenzhen Science and Technology Program (Grant No. QNXMB20250701100701002, ZDCY20250901104706008); Nanshan District Artificial Intelligence Scenario Application “Jiebang Guashuai” Project.

\bibliographystyle{elsarticle-num}
\bibliography{cas-refs}

\end{document}